\documentclass[letterpaper, 10 pt, journal]{IEEEtran}  

\IEEEoverridecommandlockouts                              

\usepackage{graphics} 
\usepackage{caption}
\usepackage{amsmath}  
\usepackage{amssymb}  
\usepackage{graphicx}
\usepackage{leftidx}
\usepackage{subfig}
\usepackage[dvipsnames]{xcolor}
\usepackage{algorithm}
\usepackage{algorithmic}
\usepackage{comment}
\usepackage{blindtext}
\usepackage[margin=1in]{geometry}
\usepackage{multirow}
\usepackage{multicol}
\usepackage{tabularray}
\usepackage{hyperref}
\usepackage{cite}
\usepackage{caption}
\usepackage{subcaption}
\usepackage{booktabs}
\usepackage[section]{placeins}
\usepackage{enumitem}

\usepackage{algorithm}
\usepackage{algorithmic}

\newcommand{\LieGroup}{\mathcal{G}}
\newcommand{\LieAlgebra}{\mathfrak{g}}
\newcommand{\AdjX}{\mathrm{Ad}_{\hat{X}_t}}
\newcommand{\ZeroMatx}{\mathbf{0}}
\newcommand{\Identity}{\mathbf{I}}
\newcommand{\dt}{\Delta t}

\usepackage{xcolor}

\usepackage{xcolor}

\title{\LARGE \bf 3D Water Quality Mapping using Invariant Extended Kalman Filtering for Underwater Robot Localization}

\author{Kaustubh Joshi$^{1}$*, Tianchen Liu$^{1}$, Alan Williams$^{2}$, Matthew Gray$^{2}$, Xiaomin Lin$^{1}$, Nikhil Chopra$^{1}$
\thanks{This work is supported by USDA NIFA Sustainable Agricultural Systems (SAS) Program (Award Number: 20206801231805).}%
\thanks{$^{1}$Maryland Robotics Center (MRC), University of Maryland, College Park, MD 20742, USA. Emails: \texttt{\{kjoshi, tianchen, xlin01, nchopra\}@umd.edu}.}
\thanks{$^{2}$Horn Point Laboratory, University of Maryland Centre for Environmental Science, Cambridge, MD 21613, USA. Emails: {\tt\small awilliams, mgray\}@umces.edu}}
\thanks{Corresponding Author: \texttt{kjoshi@umd.edu}}}

\usepackage{fancyhdr}
\fancypagestyle{withfooter}{
  
  \fancyfoot[C]{\footnotesize Accepted to the IEEE IROS workshop on Autonomous Robotic Systems in Aquaculture: Research Challenges and Industry Needs}
}

\begin{document}

\maketitle
\thispagestyle{empty}
\pagestyle{empty}

\thispagestyle{withfooter}
\pagestyle{withfooter}
\begin{abstract}

Water quality mapping for critical parameters such as temperature, salinity, and turbidity is crucial for assessing an aquaculture farm's health and yield capacity. Traditional approaches involve using boats or human divers, which are time-constrained and lack depth variability. This work presents an innovative approach to 3D water quality mapping in shallow water environments using a BlueROV2 equipped with GPS and a water quality sensor. This system allows for accurate location correction by resurfacing when errors occur. This study is being conducted at an oyster farm in the Chesapeake Bay, USA, providing a more comprehensive and precise water quality analysis in aquaculture settings.

\end{abstract}

\section{Introduction}
\label{section:introduction}

Monitoring the health of aquatic environments for aquaculture is essential for maintaining optimal water conditions to ensure the growth and productivity of shellfish and other aquaculture species. Water quality parameters such as pH, dissolved oxygen, temperature, and turbidity directly influence the health of fish and other marine organisms. Traditionally, these parameters have been monitored through point-based sampling methods, which often fail to capture the complex spatial and temporal variations in water quality, particularly in large or dynamic aquaculture environments. Existing methods include sampling using human divers, which is limited in time, or using surface vehicles, which do not provide any depth variability.

As demand for aquaculture products continues to grow, there is an increasing need for more sophisticated tools to monitor and manage water quality. Precise 3D mapping of water quality across aquaculture farms can provide a detailed understanding of how water conditions vary with depth and across different areas of the farm. This knowledge is essential to optimizing feeding strategies, minimizing disease spread, and ensuring that environmental conditions in leases remain within optimal ranges for different species.

\begin{figure}[!tp]
    \centering
    \includegraphics[width=0.8\linewidth, angle = -90]{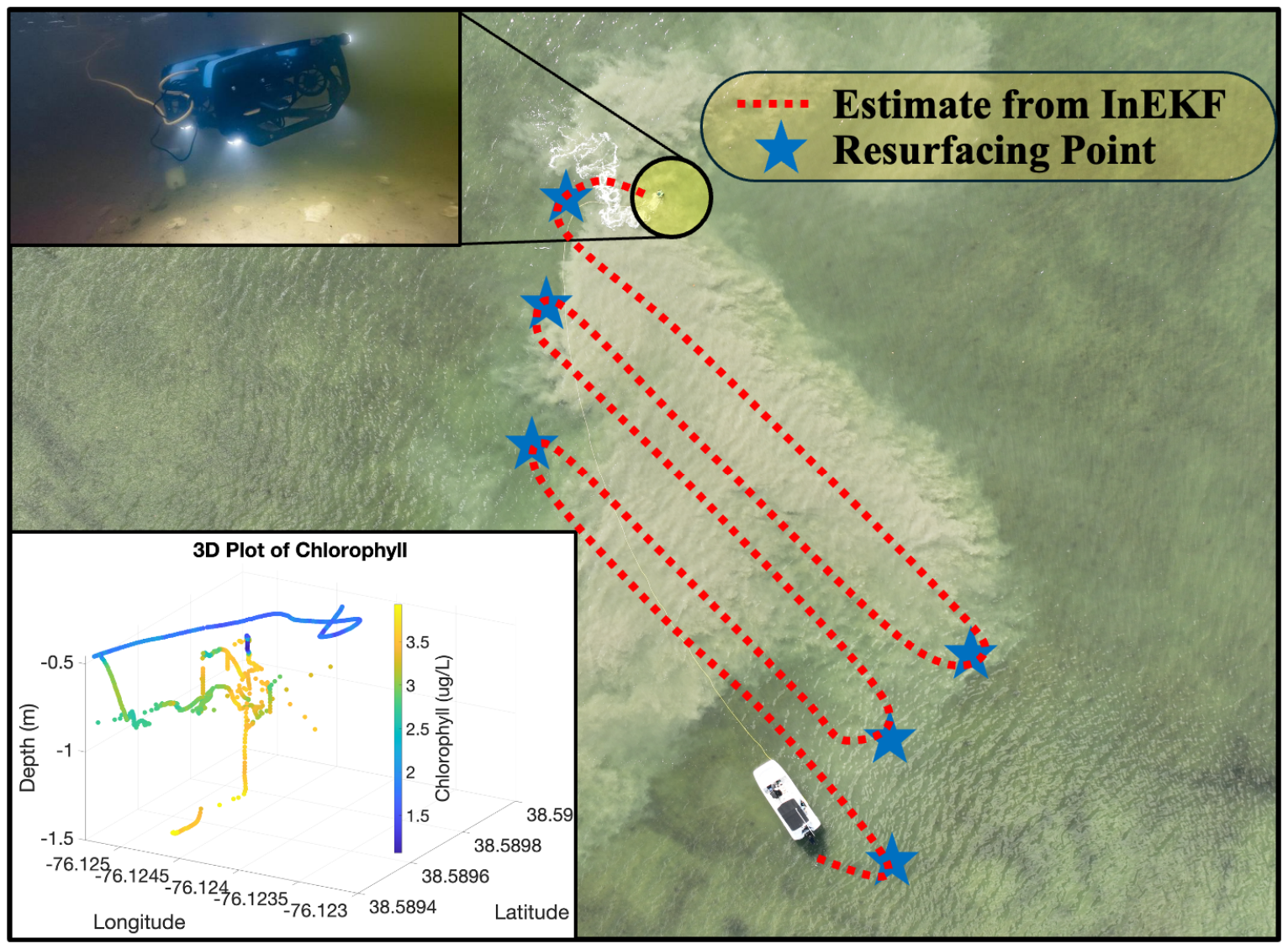}
    \caption{Concept shot of a ROV navigating a field while resurfacing at specific points. The intermediate position is estimated by the Invariant Extended Kalman Filter (InEKF). The water quality is mapped with an onboard multi-parameter sensor. A sample plot for chlorophyll concentration is shown on the bottom left.}
    \label{fig:enter-label}
\end{figure}

Recent technological advancements in underwater robotics have opened up new possibilities for continuous real-time water quality monitoring. Underwater robots with advanced sensor arrays can navigate aquaculture environments, collecting data in a wide range of depths and locations. However, the effectiveness of these systems in generating accurate 3D water quality maps depends on the precision of the robot’s localization capabilities. Underwater environments present significant challenges for localization, as traditional GPS signals do not penetrate water, requiring other sensors.

Aquaculture practices typically occur in shallow regions, making a resurfacing strategy feasible for underwater robots. However, vision-based methods and other external localization techniques are ineffective in the environments we are testing due to high water turbidity. This presents a significant challenge in obtaining accurate intermediate pose estimates between resurfacing events. We employ an Invariant Extended Kalman Filter (InEKF) \cite{barrau2015ekfslam} for state estimation to address this issue. This approach allows us to maintain reliable positioning information even when visual cues are unavailable, ensuring continuous and accurate monitoring of water quality parameters throughout the aquaculture environment.

In this study, we focus on developing a robust and accurate method for 3D water quality mapping specifically tailored to aquaculture applications. By integrating data from Doppler velocity logs (DVL), inertial measurement units (IMU), barometric pressure sensors, and periodic GPS resurfacing, our approach ensures high-precision on-board localization, critical for creating reliable 3D maps of water quality parameters. Aquaculture farmers can use these maps to make informed decisions about site management, stock distribution, and environmental interventions, ultimately leading to improved yields and sustainability in aquaculture operations.

The paper is structured as follows: Section~\ref{section:related_work} discusses work previously undertaken in the field. Then, we describe the mathematical formulation for state estimation and the process for localization in Sec.~\ref{section:methodology}.
We then present extensive quantitative evaluations of our approach in Sec.~\ref{section:Experiments_and_results}. Finally, we conclude the paper in Sec.~\ref{section:Conclusions} with parting thoughts on future work.

\section{Related Work}
\label{section:related_work}
Artificial intelligence (AI) and robotics in aquaculture have recently attracted a lot of attention. Some traditional aquaculture methods have been transformed by the use of intelligent and autonomous devices, such as unmanned surface vehicles (USVs), unmanned underwater vehicles (UUVs), and remotely operated vehicles (ROVs), which increase productivity while reducing labor costs and increasing safety. With accurate localization and navigation, these robots can help develop precise aquaculture, improving productivity. These developments are essential because the world's population increasingly relies on efficient and sustainable aquaculture methods.

\subsection{Robotics in Aquaculture}
Monitoring water quality is one of the primary uses of underwater robots in aquaculture. Conventional approaches have limited spatial coverage and are labor intensive, mostly relying on fixed point sensors and human data collecting. Several systems\cite{sun2019aquiculture, vishwas2019remotely, lin2024uivnav} have been designed to monitor the aquaculture environment’s pH value, water level, temperature, dissolved oxygen, and other water quality parameters. The advent of sophisticated sensors and robotics in recent times has made it possible to monitor water quality in real-time through the use of various robots\cite{venkatachari2024use, shen2020design,shuo2017unmanned,luna2016robotic}. These robots are highly accurate and reliable in measuring dissolved oxygen, pH, temperature, and turbidity in various water parameters over huge aquaculture areas.

In addition to environmental monitoring, robots now have built-in sophisticated feeding systems\cite{luna2016robotic,pribadi2020design,skoien2016feed} and biomass estimation tools\cite{chan2019belt,seiler2012assessing,meng2018underwater}, offering a more comprehensive approach to aquaculture management. By integrating computer vision and AI to assess biomass and monitor fish activity, these systems maximize feeding tactics while minimizing waste.

However, a number of obstacles still prevent the widespread adoption of these technologies in aquaculture, such as high deployment costs, the requirement for sturdy systems that can function in challenging marine environments, and the difficulties associated with precise localization and underwater navigation. 

\subsection{Underwater Robot Localization}
Given the critical role underwater robots play in aquaculture, particularly in environmental monitoring and autonomous feeding tasks, reliable localization is indispensable. Navigating through complicated and dynamic underwater settings and assessing the robot's status presents a substantial problem when deploying underwater robots for water quality monitoring. 

Localization for underwater vehicles is more difficult compared to other autonomous vehicles (e.g., ground or aerial vehicles) due to various environmental uncertainties. Over the years, many methods have been developed, with one prominent approach being simultaneous localization and mapping (SLAM). Visual SLAM techniques have been explored for underwater use when camera data is available, as discussed in~\cite{kim2013vslam_hull, carrasco2016stereo, weidner2017underwater}. Additionally, acoustic sensor data can enhance these systems, as demonstrated in~\cite{leutenegger2015, rahman2018sonar, vargas2021uw_vslam}. However, these methods rely heavily on visual or sonar inputs, which may not always be feasible due to limitations such as low visibility, restricted power, and limited computational resources.

Another approach involves using nonlinear observers for state estimation, with localization being a specific instance of the state estimation problem, as the vehicle's pose is part of its state. Nonlinear observers have been used for various applications, including position estimation for unmanned marine vehicles (USV and UUV)\cite{fossen1999passive}, velocity estimation\cite{liu2004nonlinear}, and wave disturbance estimation, among others. Cooperative localization has also been addressed~\cite{papadopoulos2010cooperative}. Recently, Potokar et al.~\cite{potokar2021inekf_nav} developed an InEKF-based estimation approach for underwater vehicles using IMU, DVL, and pressure sensor data, though it was only tested in simulations. In this study, we adopt a modified version of this approach to estimate the state between two resurfacing events in real-world experiments.

Building on the developments in robotics and localization, our work suggests a unique underwater robot state estimation method for 3D water quality mapping. The proposed approach aims to further develop general data-driven, intelligent aquaculture management by improving water quality measurements' geographical resolution and accuracy.

\section{Methodology}
\begin{figure}[!tp]
    \centering
    \includegraphics[width = \linewidth]{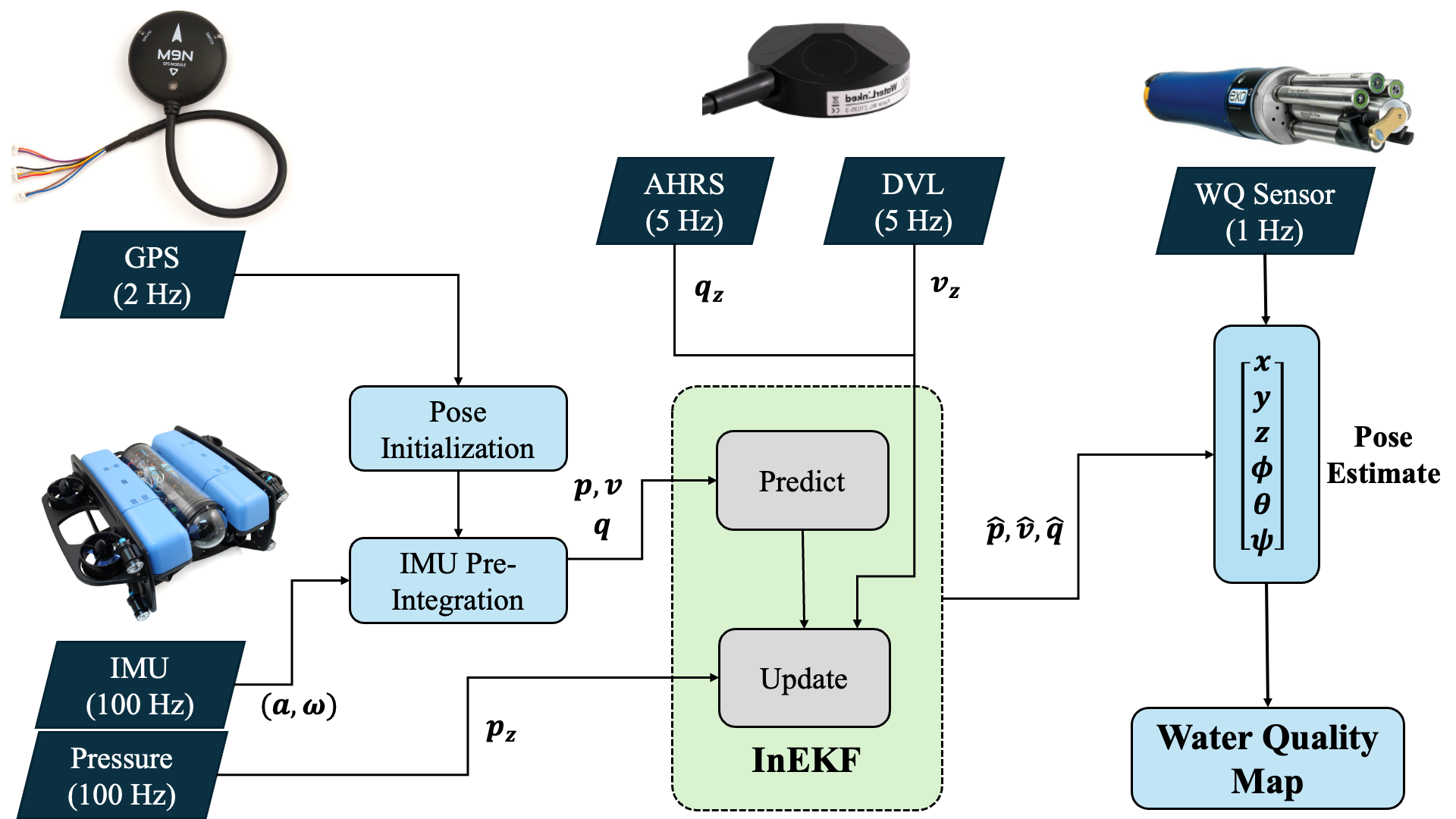}
    \caption{Process for 3D Water Quality Mapping}
    \label{fig:flowchart}
\end{figure}

The overall process is broadly accomplished in two parts: first, the pose estimation by InEKF, and second, combining the vehicle's position estimate with the water quality parameter readings. This pipeline is shown in Fig. \ref{fig:flowchart} and described in the sequel.

\label{section:methodology}

\subsection{Background and Notations}
We consider a right-invariant EKF approach~\cite{barrau2016invariant} for underwater vehicle localization based on a matrix Lie group $\LieGroup$. Specifically, the state is defined as a matrix $X \in \mathbb{R}^{5 \times 5}$ on the Lie group $\mathrm{SE}_{2}(3)$, 
\begin{equation}
	X_{t} = \left[\begin{array}{ccc}
		R_{t} & v_{t} & p_{t}  \\ 
		\ZeroMatx_{1 \times 3} & 1 & 0 \\ 
		\ZeroMatx_{1 \times 3} & 0 & 1
	\end{array}\right]
\end{equation}
where $R_{t} \in \mathrm{SO}(3)$, $v_{t} \in \mathbb{R}^{3}$ and $p_{t} \in \mathbb{R}^{3}$ are the rotation matrix, linear velocity and position of the vehicle in the world frame, respectively. The subscript $t$ refers to the variables evaluated at time $t$. Since $R_{t} \in \mathrm{SO}(3)$, the inverse of $X_{t}$ can be found as,
\begin{equation}
	X_{t}^{-1} = \left[\begin{array}{ccc}
		R_{t}^T & -R_{t}^T v_{t} & -R_{t}^T p_{t} \\ 
		\ZeroMatx_{1 \times 3} & 1 & 0 \\
		\ZeroMatx_{1 \times 3} & 0 & 1
	\end{array}\right]
\end{equation} 

For each matrix, Lie group $\LieGroup$, the associated Lie Algebra $\LieAlgebra$ is defined as the tangent space at the identity element of the group with the same dimensions. In the $\mathrm{SE}_{2}(3)$ group, an element $\chi \in \mathbb{R}^{5 \times 5}$ of $\LieAlgebra$ can be mapped to $X$ of $\LieGroup$ by applying the matrix exponential operation. On the other hand, $\chi$ can always be obtained from applying a linear map $(\cdot)^{\wedge}$ to a vector $\xi \in \mathbb{R}^{9}$ as follows,
\begin{equation}
	\chi = (\xi)^{\wedge} = \left[\begin{array}{c}
		\xi_R \\ 
		\xi_v \\ 
		\xi_p
	\end{array}\right]^{\wedge} = \left[\begin{array}{ccc}
	(\xi_R)_{\times} & \xi_v & \xi_p  \\ 
	\ZeroMatx_{1 \times 3} & 0 & 0 \\ 
	\ZeroMatx_{1 \times 3} & 0 & 0
\end{array}\right]
\end{equation}
where $\xi_R, \xi_v, \xi_p \in \mathbb{R}^{3}$. The notation $(\cdot)_{\times}$ is defined as to generate a skew-symmetric matrix in $\mathbb{R}^{3 \times 3}$ from a vector in $\mathbb{R}^{3}$, i.e.,
\begin{equation}
	(v)_{\times} = \left(\begin{array}{c}
		v_1\\
		v_2\\
		v_3
	\end{array}\right)_{\times} = \left[
	\begin{array}{ccc}
		0 & -v_3 & v_2 \\
		v_3 & 0 & -v_1 \\
		-v_2 & v_1 & 0
	\end{array}
	\right]
\end{equation}

The operation from $\mathbb{R}^{9}$ to $\LieGroup$ via $\LieAlgebra$ is completed via the matrix exponential calculation, denoted as $\textrm{exp}(\cdot)$ for brevity. 

In this paper, we use $\hat{(\cdot)}$ and $\tilde{(\cdot)}$ to represent the estimated and measured values of the variables, respectively. Following~\cite{barrau2016invariant}, the right invariant error between two trajectories $X_t$ and $\hat{X}_t$ is
\begin{equation}
	\eta_t = \hat{X}_t X_t^{-1}
\end{equation}
It is then proved that $\eta_t = \exp_{\mathrm{m}}(\xi_t^{\wedge})$ when the system satisfies certain requirements. The reader can refer to~\cite{barrau2016invariant} for more details. Then the right invariant observation is defined in the form of	
\begin{equation}
	y_t = X_t^{-1}b + W_t
    \label{eq:right_obs}
\end{equation} 
where $b$ is a known vector, $W_t$ is zero-mean Gaussian noise with covariance $Q_W$. The corresponding right invariant innovation is defined as,
\begin{equation}
	\nu_t = \hat{X}_t (y_t - \hat{y}_t)
\end{equation}
where $\hat{X}$ is the current state estimate and $\hat{y}_t$ is the measurement estimate from $\hat{X}$. 

By the first-order approximation $ \eta_t = \exp_{\mathrm{m}}(\xi_t^{\wedge}) \approx \Identity + \xi_t^{\wedge}$, the innovation can be rewritten as,
\begin{align}
	\nu_t & = \hat{X}_t (y_t - \hat{y}_t) = \hat{X}_t (X_t^{-1}b + W_t - \hat{X}_t^{-1}b) \nonumber \\
	& = \hat{X}_t X_t^{-1}b + \hat{X}_t W_t - b = \eta_t b + \hat{X}_t W_t - b \nonumber \\
	& \approx (\Identity_5 + \xi_t^{\wedge})b + \hat{X}_t W_t - b = \xi_t^{\wedge}b + \hat{X}_t W_t
\end{align}
where $\Identity_{N}$ denotes the identity matrix of dimension $N \times N$. A matrix $H$ is defined that satisfies
\begin{equation}
	H\xi_t^{\wedge} = -\xi_t^{\wedge}b
	\label{eq: H_matrix_eq}
\end{equation}
Given a constant $b$ vector, $H$ can be found and then used in the update steps. 

\subsection{Proposed Approach Description}
The proposed approach utilizes multiple sensors commonly available in underwater vehicles to fulfill the localization task, namely IMU, DVL, and a pressure sensor. The IMU sensor obtains high-frequency linear acceleration and angular velocity measurements in autonomous systems. Hence, in theory, dead-reckoning from IMU measurements alone can provide accurate localization results. However, due to hardware issues, the IMU measurements are not always precise, and other sensor measurements are necessary to correct incorrect estimates. The approach is described in Algorithm~\ref{al:inekf} in general, and the details are presented in the following subsections. For conciseness of notation, the subscripts `$t+1$' and `$t+1|t$' of the state and covariance to denote the corrections are dropped in the update steps.   

\begin{algorithm}
	\caption{InEKF Localization Approach}\label{al:inekf}
	\begin{algorithmic}
		\STATE State: $X = \left[\begin{array}{ccc}
			R & v & p  \\ 
			\ZeroMatx_{1 \times 3} & 1 & 0 \\ 
			\ZeroMatx_{1 \times 3} & 0 & 1
		\end{array}\right], R \in SO(3), v, p \in \mathbb{R}^{3}$, Covariance: $P =  0.1 \, \Identity_{9}$
		\FOR{\text{all timestamps}}
		\STATE \textbf{Prediction}
		\STATE \quad Calculate $\hat{X}, \hat{P}$ using Eq.~\eqref{eq:imu_start} - Eq.~\eqref{eq:imu_end}
		\STATE \textbf{Update from DVL}
		\STATE \quad Calculate $\hat{X}, \hat{P}$ using Eq.~\eqref{eq:dvl_start} - Eq.~\eqref{eq:dvl_end}
		\STATE \textbf{Update from Pressure Sensor}
		\STATE \quad Calculate $\hat{X}, \hat{P}$ using Eq.~\eqref{eq:depth_start} - Eq.~\eqref{eq:depth_end}
		\ENDFOR
	\end{algorithmic}
\end{algorithm}

\subsection{IMU Motion Model}
First, the IMU sensor data is used to predict the current state. As mentioned previously, the measurements are usually not precise due to sensor noise, so we assume that zero-mean Gaussian noises offset the actual measurements,
\begin{align}
	\tilde{\omega}_t & = \omega_t + w_t^{g}, \quad \tilde{a}_t = a_t + w_t^{a} 
\end{align}
where $\tilde{\omega}_t$ and $\tilde{a}_t$ are measurements of the angular velocity and the linear acceleration. $w_t^{g} \sim \mathcal{N}(\ZeroMatx_{3 \times 1}, Q_{\omega})$ and $w_t^{a} \sim \mathcal{N}(\ZeroMatx_{3 \times 1}, Q_{a})$ are the Gaussian noises of the gyroscope and the accelerator, respectively. 

The IMU dynamic model has been proven to satisfy the requirements for the InEKF approach (e.g.~\cite{zhang2017convergence,hartley2020contact}). As high-frequency sensor data, it can be assumed that IMU readings are constant during the short time interval. The effect of the angular velocity on the acceleration is also assumed to be sufficiently small, so this is neglected. Hence, the following discrete dynamic equations are used for the prediction step.  
\begin{align}
	\hat{R}_{t+1|t} & = \hat{R}_t \exp((\tilde{\omega}_t)_{\times} \dt) \label{eq:imu_start} \\
	\hat{v}_{t+1|t} & = \hat{v}_t + (\hat{R}_t (\tilde{a}_t) + g) \dt \\
	\hat{p}_{t+1|t} & = \hat{p}_t + \hat{v}_t \dt + \frac{1}{2} (\hat{R}_t (\tilde{a}_t) + g) \dt^2
\end{align}

The state covariance $\hat{P}$ is updated by
\begin{equation}
	\hat{P}_{t+1|t} = \left(\Phi \hat{P}_t \Phi^T + \Phi \AdjX Q \AdjX^T \Phi^T \right) \dt
\end{equation}
where $Q = \mathrm{block{\_}diag}(Q_{\omega}, Q_{a}, \ZeroMatx_{3 \times 3})$ is the covariance matrix of the noises and
\begin{align}
	\Phi & = \exp \bigg(
	\left[\begin{array}{ccc}
		\ZeroMatx_{3 \times 3}  & \ZeroMatx_{3 \times 3}  & \ZeroMatx_{3 \times 3}  \\ 
		(g)_{\times} & \ZeroMatx_{3 \times 3}  & \ZeroMatx_{3 \times 3}  \\
		\ZeroMatx_{3 \times 3} & \Identity_{3} & \ZeroMatx_{3 \times 3} 
	\end{array}\right]  \dt \bigg)	\label{eq:imu_end}
\end{align}

Following the adjoint map definition in~\cite{hall2003lie}, the matrix $\AdjX$ is defined such that $\AdjX(\xi^{\wedge}) = (\AdjX \xi)^{\wedge}$ is satisfied, hence
\begin{equation}
	\AdjX = \left[\begin{array}{ccc}
		R_{t} & \ZeroMatx_{3 \times 3} & \ZeroMatx_{3 \times 3}  \\ 
		(v_{t})_{\times} R_{t} & R_{t} & \ZeroMatx_{3 \times 3} \\ 
		(p_{t})_{\times} R_{t} & \ZeroMatx_{3 \times 3} & R_{t}
	\end{array}\right] 
\end{equation}

Note that the above equations are valid when the IMU frame is set as the vehicle's body frame. Also, the IMU bias is not discussed in this paper, but it can be included using an `imperfect InEKF' as shown in~\cite{hartley2020contact, potokar2021inekf_nav}.

\subsection{Update from DVL Measurement}
In the underwater environment, a DVL is used to estimate the vehicle's speed by sending and receiving acoustic waves. We follow~\cite{potokar2021inekf_nav} to incorporate DVL sensor data in the approach. It is assumed that the observed velocity values $\tilde{y}_{{D}} \in \mathbb{R}^{3}$ are offset by a zero-mean white Gaussian noise. If the rigid body transformation between the IMU frame and the DVL frame is denoted by rotation matrix $R_{DI}$ and translation vector $t_{DI}$, the measured vehicle velocity $\tilde{y}_{{D}}$ is then transformed into IMU frame as $\tilde{v}$,
\begin{align}
	\tilde{v} & = R_{DI} \tilde{y}_{{D}} + (t_{DI})_{\times} (\tilde{\omega}) \label{eq:dvl_start}
\end{align}
The covariance of $\tilde{v}_{t}$ after transformation is $Q_v = R_{DI} Q_D R_{DI}^T + (t_{DI})_{\times} Q_{\omega} (t_{DI})_{\times}^T$, where $Q_D$ is the covariance matrix of DVL noise. By defining $b_{D} = \left[0, 0, 0, 1, 0 \right]^T$, $H_D$ can be found by Eq.~\eqref{eq: H_matrix_eq} as 
\begin{equation}
	H_D = \left[\begin{array}{ccc}
		\ZeroMatx_{3 \times 3}  & -\Identity_{3}  & \ZeroMatx_{3 \times 3} \\
		\ZeroMatx_{2 \times 3}  & \ZeroMatx_{2 \times 3}  & \ZeroMatx_{2 \times 3}
	\end{array}\right]
\end{equation}
The observation is reset as $z = \left[\tilde{v}_{1}, \tilde{v}_{2}, \tilde{v}_{3}, 1, 0 \right]^T$ for obtaining the proper dimensions for matrix multiplication. Then, the conventional Kalman theory is applied to correct the prediction state $\hat{X}$ and covariance $\hat{P}$ as follows,
\begin{align}
	S & = H_D \hat{P} H_D^T + \hat{X} \bar{Q}_v \hat{X}^T, \quad K = \hat{P} H_D^T S^{-1} \\
	\hat{X} & = \exp ((K \hat{X} z )^{\wedge}) \hat{X}, \quad \hat{P} = (\Identity_{9} - K H_D) \hat{P} \label{eq:dvl_end}
\end{align}
where $\bar{Q}_v = \mathrm{block{\_}diag}(Q_{v}, \ZeroMatx_{2 \times 2})$.

\subsection{Update from Pressure Measurement}
The pressure sensor is widely used to measure the depth of an underwater vehicle. Due to the easy conversion from the water pressure, the depth can be obtained with high resolution as an accurate way to localize the vehicle. The observed value $\tilde{y}_{P} \in \mathbb{R}$ of the pressure sensor is approximately proportional to the real depth ${p}_{z}$. Without loss of generality, we use $\tilde{y}_{depth}$ to represent $\tilde{y}_{P}$ to denote the measured depth.

In~\cite{potokar2021inekf_nav}, $\tilde{y}_{depth}$ is accompanied with `pseudo' measurements $\hat{p}_x$ and $\hat{p}_y$ to form a left-invariant observation. We use the same idea to create the `measurement' of the vehicle position, defined as $\tilde{p} = \left[\hat{p}_x, \hat{p}_y, \tilde{y}_{depth}\right]^T$. But instead of estimating $\hat{X}$ and $\hat{P}$ in a left-invariant setup, we directly work with the right-invariant observation and innovation. Defining $b_{depth} = \left[0, 0, 0, 0, 1 \right]^T$, $H_P$ can be found by Eq.~\eqref{eq: H_matrix_eq} as,
\begin{equation}
	H_P = \left[\begin{array}{ccc}
		\ZeroMatx_{3 \times 3} & \ZeroMatx_{3 \times 3} & -\Identity_{3} \\
		\ZeroMatx_{2 \times 3} & \ZeroMatx_{2 \times 3} & \ZeroMatx_{2 \times 3}
	\end{array}\right]
\end{equation}
To directly work in the right invariant setup, the `pseudo' measurements must be converted using the right observation equation (Eqn. (\ref{eq:right_obs})), i.e., $\tilde{p} = -\hat{R}^T \tilde{p}$. This observation is reset as $\tilde{y}_{depth} = \left[\tilde{p}_1, \tilde{p}_2, \tilde{p}_3, 0, 1 \right]^T$ for proper dimension. Then the innovation can be written as,
\begin{equation}
	\nu = \hat{X} \tilde{y}_{depth} - b_{depth} \label{eq:depth_start} 
\end{equation}
After this conversion, $\hat{X}$ and $\hat{P}$ can be updated using the depth measurement following a standard right InEKF update step~\cite{barrau2016invariant},
\begin{align}
	S & = H_P \hat{P} H_P^T + Q_P, \quad K = \hat{P} H_P^T S^{-1} \\ 
	\hat{X} & = \hat{X} \exp ((K \nu)^{\wedge}), \quad \hat{P} = (\Identity_{9} - K H_P) \hat{P} \label{eq:depth_end}
\end{align}
where $\bar{Q}_P = \mathrm{block{\_}diag}(Q_{P}, \ZeroMatx_{2 \times 2})$ and $Q_{P}$ is the covariance matrix of `pseudo' measurement noises.

Using the process shown in Algorithm \ref{al:inekf}, we carry out the intermediate state estimation for the robot between two GPS resurfacing.







\section{Setup \& Experimental Results}
\label{section:Experiments_and_results}
To check its efficacy, we validate and test InEKF on data collected with our system.



\subsection{Experiments with BlueROV2}

\begin{figure}
    \centering
    \includegraphics[width=\linewidth]{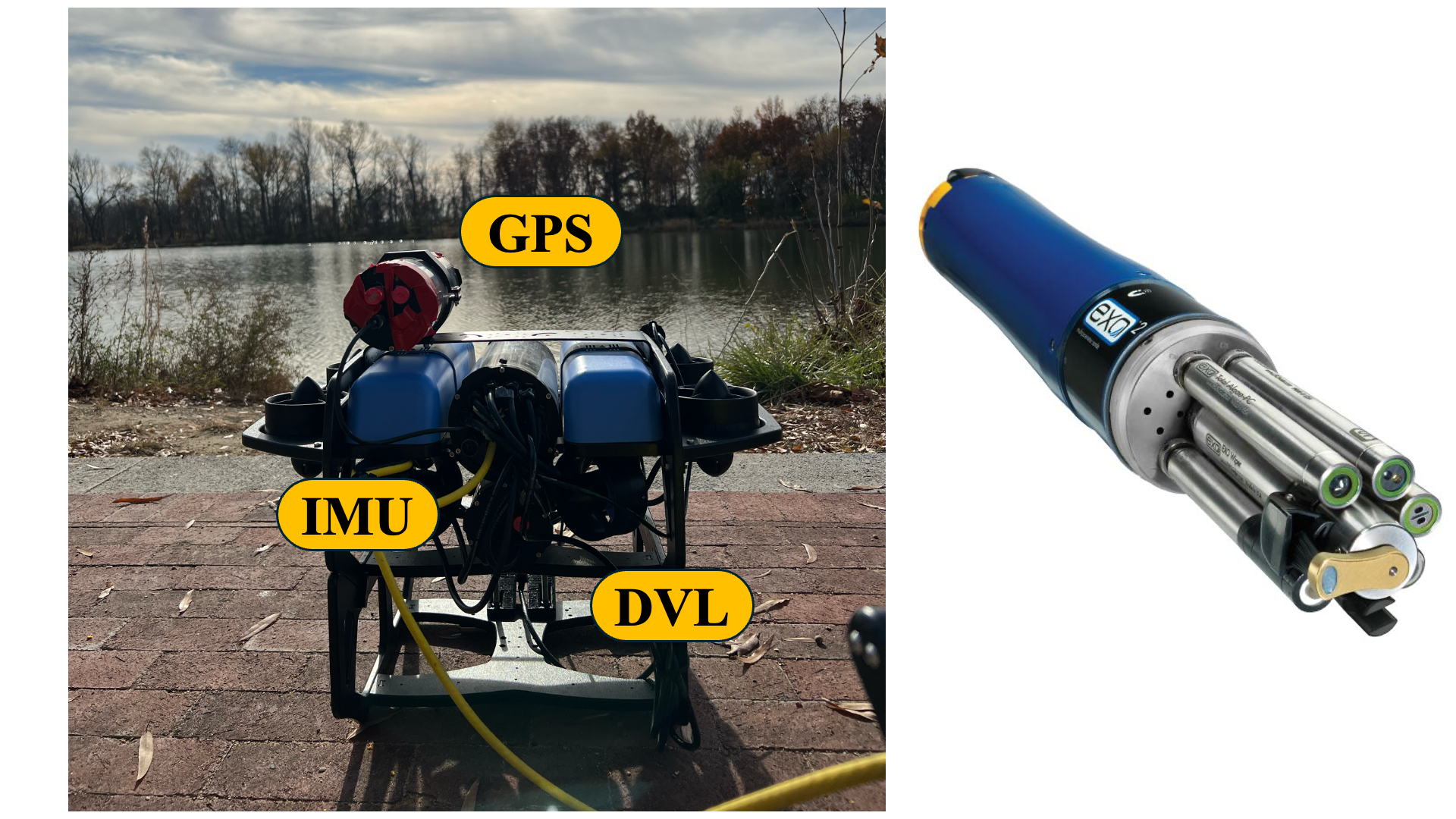}
    \caption{\textit{(left)} The BlueROV2 platform with retrofitted GPS mount and DVL. \textit{(right)} The EXO2 multiparameter sensor is attached to the ROV for water quality analysis.}
    \label{fig:system}
\end{figure}

\begin{figure}
    \centering
    \includegraphics[width=\linewidth]{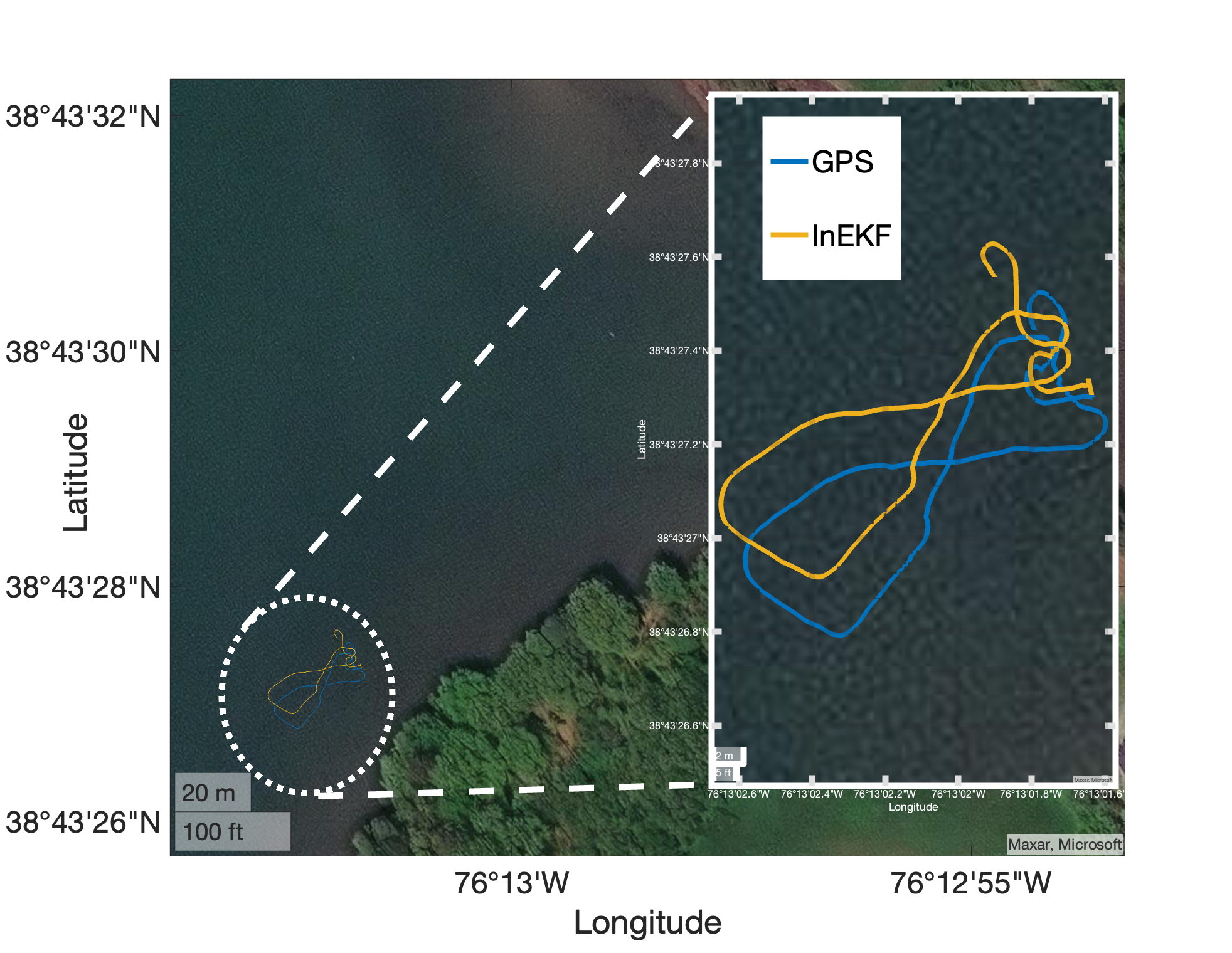}
    \caption{Comparison of Surface GPS reading vs InEKF estimate. The robot was initialized with GPS reading at the start point.}
    \label{fig:uuv-inekf}
\end{figure}

\begin{table}
    \centering 
    \caption{Comparison of MAE in individual states, overall RMSE, and total runtime for EKF and InEKF for different time intervals}
    \begin{tabular}{lcccl}\toprule
MAE for State & \multicolumn{2}{c}{BlueROV2 (100 s)} & \multicolumn{2}{c}{BlueROV2 (300 s)}
\\\cmidrule(lr){2-3}\cmidrule(lr){4-5}
          &  EKF  & InEKF  & EKF  & InEKF \\\midrule
$\mathbf{x}$  & 0.7698 & 0.4812 & 1.5410 & 1.2766 \\
$\mathbf{y}$  & 0.8997 & 1.2794 & 1.4022 & 2.6352\\
$\mathbf{z}$  & 0.1967 & 0.0631 & 0.3060 & 0.0758\\ \midrule
Total Error  & 5.7409 & \textbf{1.5065} & 8.0703 & \textbf{3.0448}\\
Total Variance  & 7.6428 & \textbf{0.6042} & 26.0001 & \textbf{2.0545}\\\midrule
Runtime (in s)  & \textbf{3.0306} & 11.0643 & \textbf{9.0348} & 19.5921\\\bottomrule
\end{tabular}
\label{tab:results}
\end{table}

\begin{figure*}
    \centering
    
    \begin{minipage}[b]{0.3\textwidth}
        \centering
        \includegraphics[width=\textwidth]{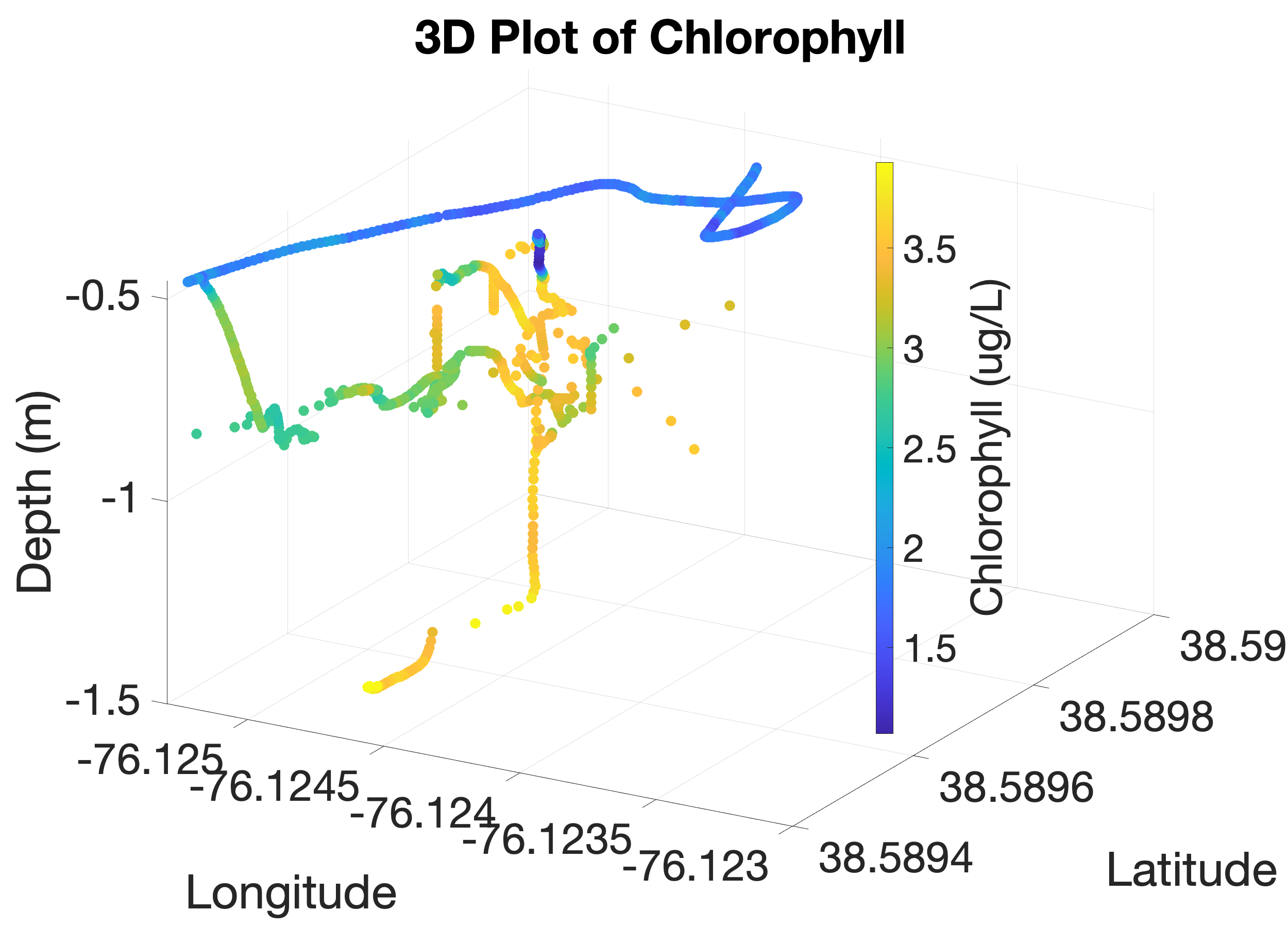}
        \caption{Chlorophyll}
        \label{fig:chlorophyll}
    \end{minipage}
    \hfill
    \begin{minipage}[b]{0.3\textwidth}
        \centering
        \includegraphics[width=\textwidth]{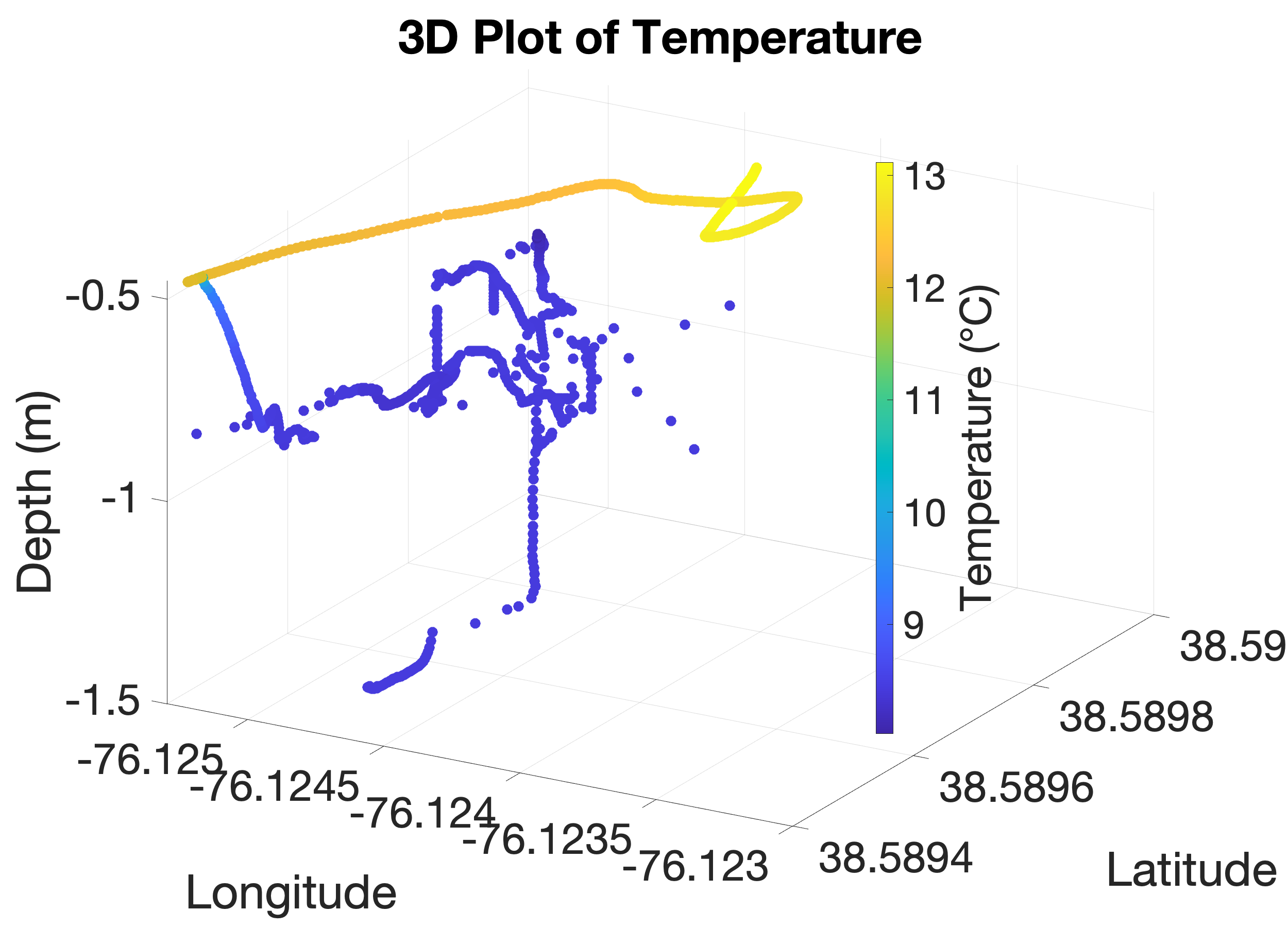}
        \caption{Temperature}
        \label{fig:temperature}
    \end{minipage}
    \hfill
    \begin{minipage}[b]{0.3\textwidth}
        \centering
        \includegraphics[width=\textwidth]{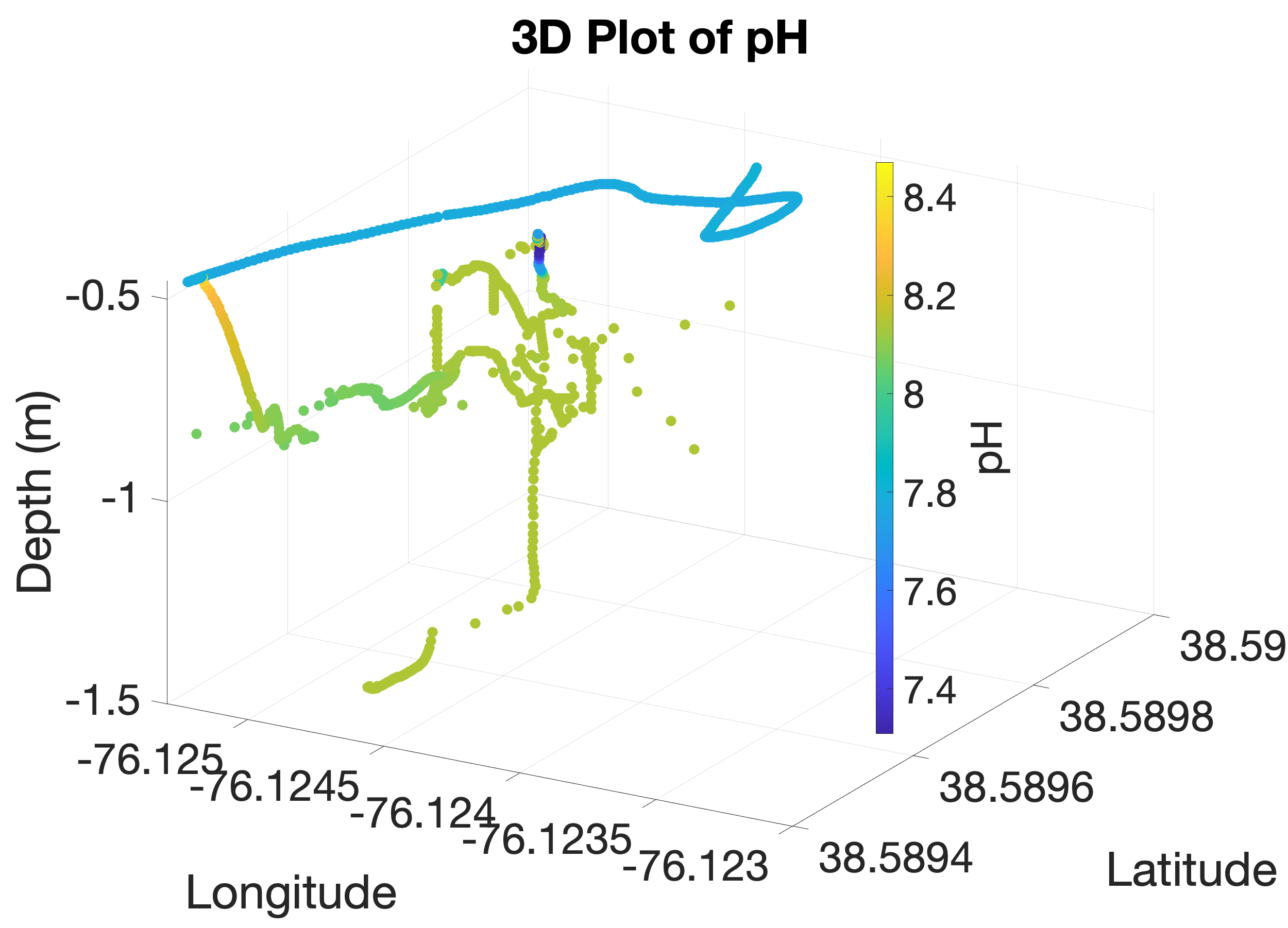}
        \caption{pH}
        \label{fig:ph}
    \end{minipage}
    
    \vspace{0.2cm} 
    
    \begin{minipage}[b]{0.3\textwidth}
        \centering
        \includegraphics[width=\textwidth]{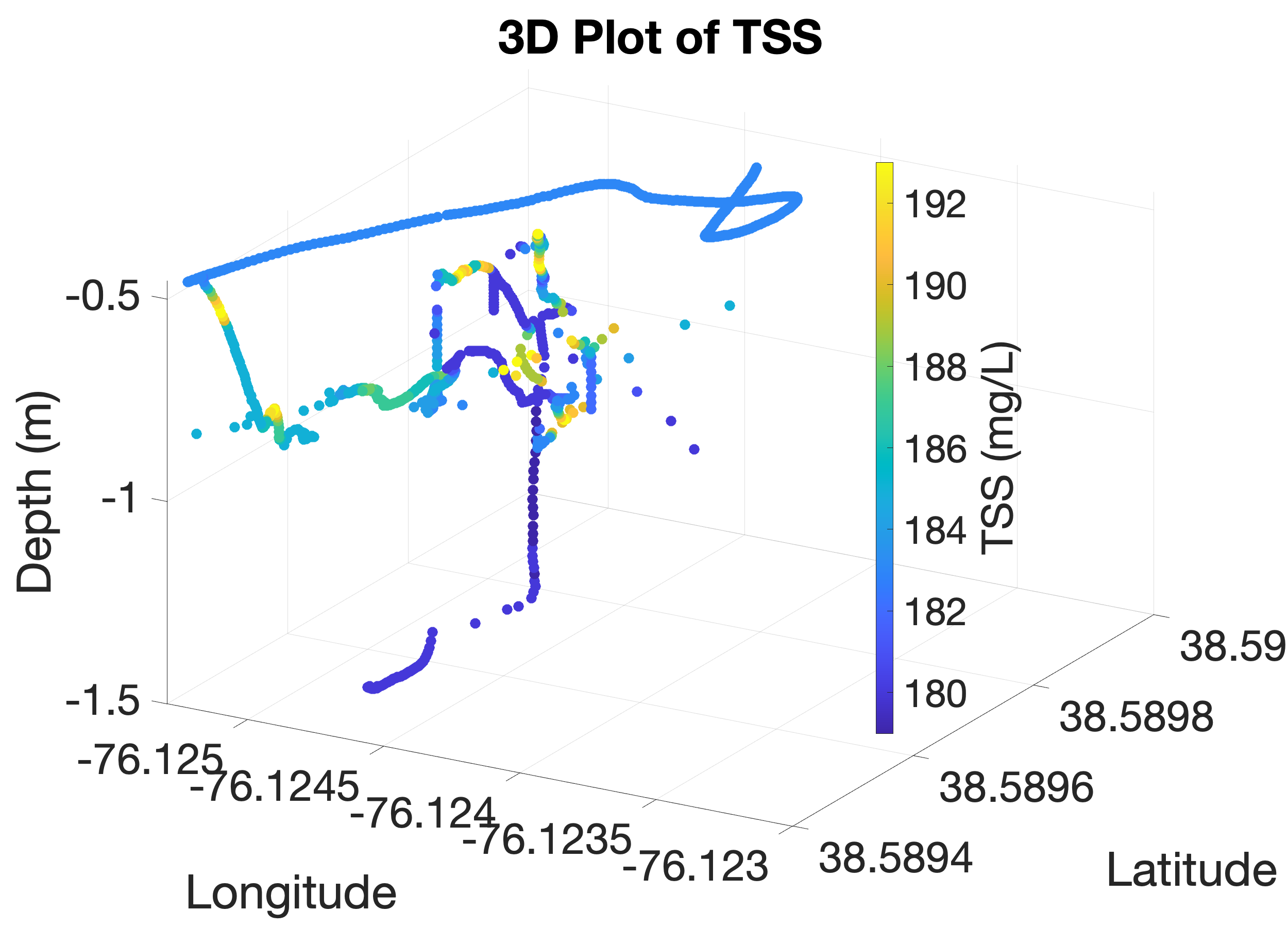}
        \caption{Total Suspended Solids (TSS)}
        \label{fig:tss}
    \end{minipage}
    \hfill
    \begin{minipage}[b]{0.3\textwidth}
        \centering
        \includegraphics[width=\textwidth]{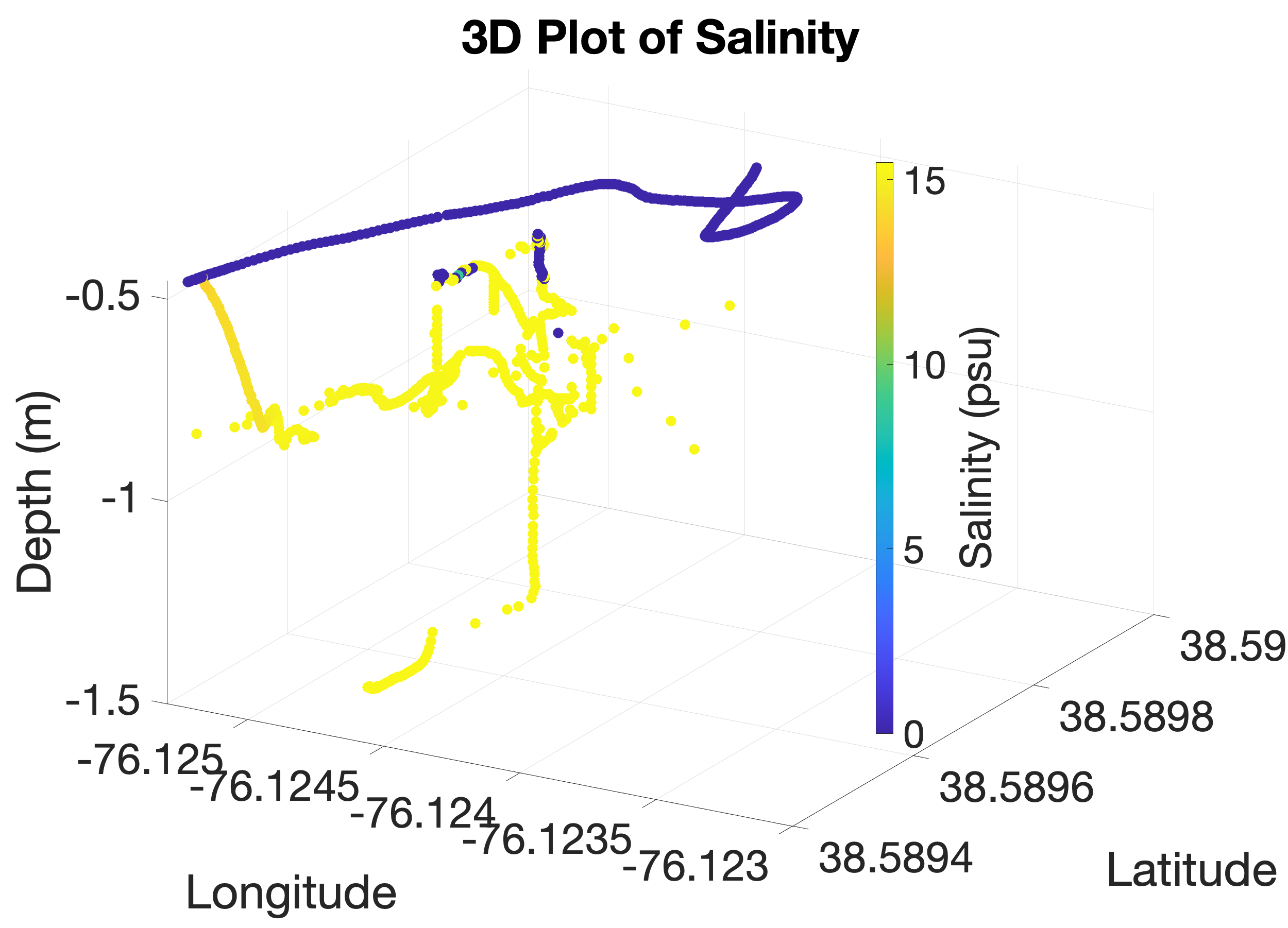}
        \caption{Salinity}
        \label{fig:salinity}
    \end{minipage}
    \hfill
    \begin{minipage}[b]{0.3\textwidth}
        \centering
        \includegraphics[width=\textwidth]{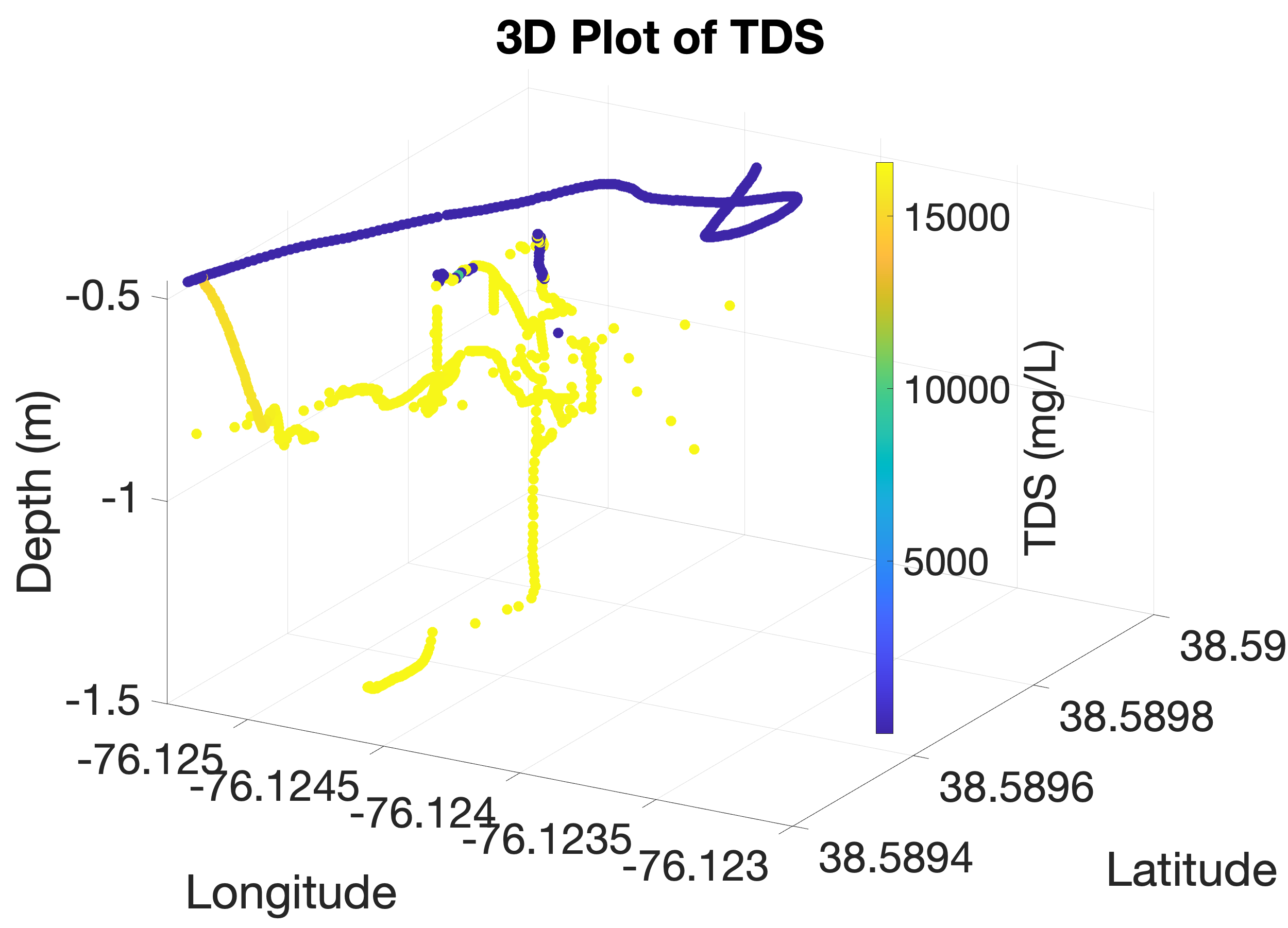}
        \caption{Total Dissolved Solids (TDS)}
        \label{fig:tds}
    \end{minipage}
    
    \label{fig:3x2grid}
\end{figure*}

Our real-world experiments take place in multiple locations within the Chesapeake Bay, designated for oyster leases. The aim of these experiments is to track water quality variability using sensors mounted on the BlueROV2 platform, where accurate localization of the vehicle is essential. The system setup and sensors are shown in Fig. \ref{fig:system} and described below.

The BlueROV2 Heavy Configuration \cite{BlueROV2} is employed for the experiments, featuring 8 T200 thrusters that enable motion across all 6 degrees of freedom. The vehicle is equipped with a TDK InvenSense ICM-20602 (IMU) running at 100Hz, a Waterlinked DVL-A50 (DVL) with built-in AHRS operating at 20Hz, and a u-blox M9N GPS module functioning at 5Hz. Control of the ROV is achieved through Pulse-Width Modulation (PWM) signals sent to the thrusters. The GPS sensor, mounted half meter above the ROV’s frame, provides ground truth measurements. The system’s affordability allows validation of the InEKF approach using low-cost sensors. The ROV operates just below the water surface at approximately 0.2m, fully submerged while keeping the GPS above the surface to maintain continuous signal reception. The vehicle was operated at low speeds (less than 1 m/s) on various trajectories in the bay, with all sensor data collected.

The datasets obtained from the Unmanned Underwater Vehicle (UUV) were analyzed using both the Invariant Extended Kalman Filter (InEKF) and the traditional Extended Kalman Filter (EKF). For both filters, the covariance matrices were set to $P = 0.1 \mathbf{I}$, $Q = 0.1 \mathbf{I}$, and $R = 0.1 \mathbf{I}$. These algorithms were developed and tested on a MacBook Pro equipped with an M1 processor using Python scripts.
The resulting trajectories were overlaid on a satellite image in Fig. \ref{fig:uuv-inekf}, with coordinates converted from the Mercator projection \cite{snyder1987map} to geographic coordinates. While InEKF demonstrated significantly better accuracy than EKF, it required more computational time. However, the processing time remained shorter than the total trajectory duration, indicating that InEKF is feasible for real-time application. The Mean Absolute Error (MAE) for individual states and the overall Root Mean Squared Error (RMSE) are summarized in Table \ref{tab:results}, along with the runtime for each method. The table provides estimates for short-term (100s) and medium-term (300s) performance. Short-term estimates suggest the platform could still function during brief sensor outages, with total error remaining around 1.5 meters. In the case of longer outages, onboard inertial sensors can continue to provide relatively accurate state estimates, even on a low-cost system.

\subsection{Water Quality Mapping}

The EXO2 is a multiparameter sensor that was used for analyzing the water quality. Due to its heavy weight, changes were made to the robot to ensure neutral buoyancy for maximum energy efficiency. The sensor was attached to the ROV frame and initialized at the beginning of the ROV deployment to record data. This data was analyzed post-experiments, and the timestamps were synchronized with the GPS timestamps for 1Hz frequency. 3D mappings of multiple parameters like temperature, chlorophyll, pH, salinity, turbidity, etc. were created. The results for one of the tests in the Chesapeake Bay over a bottom culture oyster lease are shown in the Fig.~\ref{fig:chlorophyll} -- \ref{fig:tds}. The results demonstrate how an unmanned underwater vehicle with an accurate and robust localization technology can assist in generating 3D maps of the water quality of a shallow water body. 

\section{Conclusion and Future Work}
\label{section:Conclusions}
In this paper, preliminary studies have been undertaken to demonstrate how an underwater localization method, i.e., InEKF, can map the water quality of a shallow water body. We presented the theoretical approach of InEKF and tuned it for an underwater robot system. One significant advantage of this approach is that it is agnostic to any robot platform since it is independent of the system model and only works based on sensor data. Next, we tested the algorithm in the field using our system, obtaining better results than the commonly used EKF algorithm for state estimation. Finally, we attached a multiparameter water quality sensor to our system and obtained a 3D water quality map for an oyster lease.

\textit{Future Work: } In our tests, InEKF guaranteed convergence, but it did not always obtain the best accuracy. Hence, new state estimation methods can be devised to provide a more accurate result. Another improvement for the underwater vehicle can be the real-time integration of a multiparameter sensor. As the current WQ sensor is integrated mechanically, a real-time connection to the vehicle can enable autonomous navigation to areas of interest with specific parameter values.


\section*{ACKNOWLEDGMENTS}
The authors extend their gratitude to Dr. Don Webster and Mr. Bobby Leonard for providing access to oyster leases, and to Logan Bilbrough for the aerial imagery.

\bibliographystyle{IEEEtran}
\bibliography{refs_ipg_observer, refs_uuv, refs_mhe, refs_rov, refs_imu,refs_aqual}

\end{document}